\documentclass[letterpaper]{article}
\usepackage{aaai2027}
\usepackage[hyphens]{url}
\usepackage{graphicx}
\usepackage{natbib}
\usepackage{caption}
\usepackage{newfloat}
\usepackage{listings}
\DeclareCaptionStyle{ruled}{labelfont=normalfont,labelsep=colon,strut=off}
\DeclareFloatingEnvironment[name=Listing,placement=tb,fileext=lst]{listing}
\usepackage{booktabs}
\nocopyright

\title{Wactorz: Self-Writing LLM Agents That Discover, Repair, and Migrate at the Edge\\
(Extended Version)}
\author{
  Panagiotis Kasnesis\textsuperscript{\rm 1},
  Lazaros Toumanidis\textsuperscript{\rm 1},
  Amalia Contiero\textsuperscript{\rm 2},
  Christos Chatzigeorgiou\textsuperscript{\rm 1}
}
\affiliations{
  \textsuperscript{\rm 1}Waldiez PC, Maroussi, Greece\\
  \textsuperscript{\rm 2}ThinGenious PC, Maroussi, Greece\\
  \{pkasnesis, laztoum, chrichatz\}@waldiez.io, amaliacontiero@gmail.com
}

\begin{document}
\maketitle

\begin{abstract}
Most LLM agent frameworks build a crew that runs a task and exits. Physical
environments (e.g., homes, labs, factory floors) need agents that stay up for
months, survive their own bugs, and run on whatever hardware is already next to
the sensors. We present \textbf{Wactorz}, an open-source runtime in which
LLM-generated agents are long-lived, supervised \emph{actors}: a planner turns
a chat request into Python, compiles it into a new actor at runtime, and a
supervisor restarts that actor alone, with its state intact, when it crashes. All
coordination rides on MQTT, so agents can be spawned on, and live-migrated
between, edge nodes such as a Raspberry~Pi, and every internal event is a
topic external systems can subscribe to. The recorded demo builds a camera-to-lamp automation from one chat
request, adds a second agent that is auto-wired to the first by topic and
carries a deliberate division-by-zero, watches
that agent repair its own code without losing its count, then deploys a fresh
Raspberry~Pi node and live-migrates the agent onto it, in under four minutes of
wall-clock time and \$0.13 of model usage. This extended version adds what the
two-page demonstration paper could not: the Wiring view of the topic graph,
Wactorz as a library for hand-written agents, how trained models are stored,
sent to nodes and replaced, the security model, and the system's limitations.
\end{abstract}

\begin{links}
\link{Code}{https://github.com/waldiez/wactorz}
\end{links}

\section{Introduction}

LLM orchestration libraries such as AutoGen~\cite{wu2023autogen},
CrewAI and LangGraph~\cite{chase2022langchain} model a multi-agent system as
a conversation or graph that is executed inside one script and terminates when
the task does. That is the wrong lifecycle for \emph{physical AI}: an agent that
watches a camera feed, a power meter, or a door sensor must run continuously,
tolerate faults in code it did not write, and often has to sit on a
low-power device with intermittent cloud access. Existing edge automation
tools (Home Assistant rules, Node-RED flows) are long-lived but static: a human
writes every rule.

Wactorz closes that gap by combining three ideas from different communities:
the \emph{actor model} with supervision trees from the Erlang/OTP tradition
\cite{hewitt1973actor,armstrong2003thesis}; \emph{code-generating LLM agents}
that write their own behaviour at runtime \cite{wang2023voyager}; and a \emph{pub/sub IoT bus} (MQTT) as the sole coordination substrate
\cite{mqtt5}, on which agents discover one another by observed topic rather
than by the ontology-driven matching of needs and services we proposed
for the social IoT \cite{kasnesis2017siot}. The result is a system in which the LLM is the programmer, the
supervisor is the safety net, and the broker is the network. To our
knowledge it is the first agent framework in which LLM-written agents are
first-class, supervised, migratable actors.

The same runtime serves agents nobody generated. A developer can hand Wactorz
a Python function and get the same supervision, persistence, wiring and
migration, and a trained model can be stored once on the main machine and
served, by version, to whichever node runs the agent that needs it. This
version describes Wactorz~0.8.

\section{System Description}

\begin{figure}[t]
\centering
\includegraphics[width=\columnwidth]{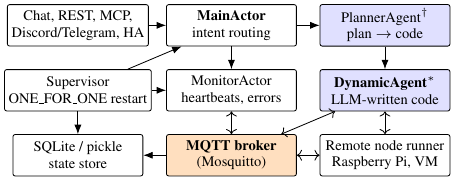}
\caption{Wactorz architecture. Blue nodes are created at runtime:
$^\dagger$a PlannerAgent per request, which self-terminates, and $^*$the
DynamicAgents it writes, re-spawned from the persisted registry on restart.}
\label{fig:arch}
\end{figure}

Wactorz is a single-process Python \texttt{asyncio} actor system
(Figure~\ref{fig:arch}); a separate Mosquitto broker is the only other
component. Every agent is an \texttt{Actor} with an inbox, a message loop, a
heartbeat every 10\,s, and a key-value \texttt{persist()}/\allowbreak\texttt{recall()} API that
routes structured keys to SQLite, trained models, \texttt{numpy} arrays and
tensors to blob files of their own, and other Python objects to pickle.
Each agent has an identity that survives restarts and moves, and is
addressed as \texttt{name@node}, so two machines may each run an agent of
the same name.

\subsection{Agents that write agents}
A user types a request in natural language, e.g.\ \emph{``when a person is
detected on my PC camera, turn on the office light''}. \texttt{MainActor}
classifies the intent (direct actuation, Home Assistant query, pipeline, or
chat). Pipelines go to an ephemeral per-request \texttt{PlannerAgent}, which emits a JSON
plan and, for each step, Python source defining \texttt{setup()},
\texttt{process()} and \texttt{handle\_task()}. Each block becomes a
\texttt{DynamicAgent}, spawned into the running system with no restart. An \texttt{Installer\-Agent} \texttt{pip}-installs declared
dependencies before a spawn; a package that is not on an allow-list waits
until the operator approves it on the dashboard. A catalogue of trusted
recipes (time-series collection, anomaly detection, weather, energy) covers
common needs without generated code.

\subsection{Discovery by topic, not by name}
Agents never address each other by name. Each publishes and subscribes through
MQTT topics, and a \emph{TopicBus} registers a \emph{contract} per topic, including its
producer, consumers and a captured sample of the real payload. When a later
request needs data that already flows (``count the people the camera
sees''), the planner is handed the matching contracts and samples before it
writes code, so the consumer reads \texttt{payload["person"]} because that is
what the producer actually sends, not what a second LLM call guessed. This
removes the classic failure of multi-LLM pipelines, two agents disagreeing on
field names, and makes agents replaceable: rename or regenerate a producer and
its consumers keep working as long as the topic and shape hold. Our earlier
social-IoT work matched objects' needs to services through ontologies and
semantic rules \cite{kasnesis2017siot}; contracts learned from live payloads
need no shared vocabulary in advance, which is what agents whose schemas do
not exist until an LLM writes them require.

\begin{figure*}[t]
\centering
\includegraphics[width=\textwidth]{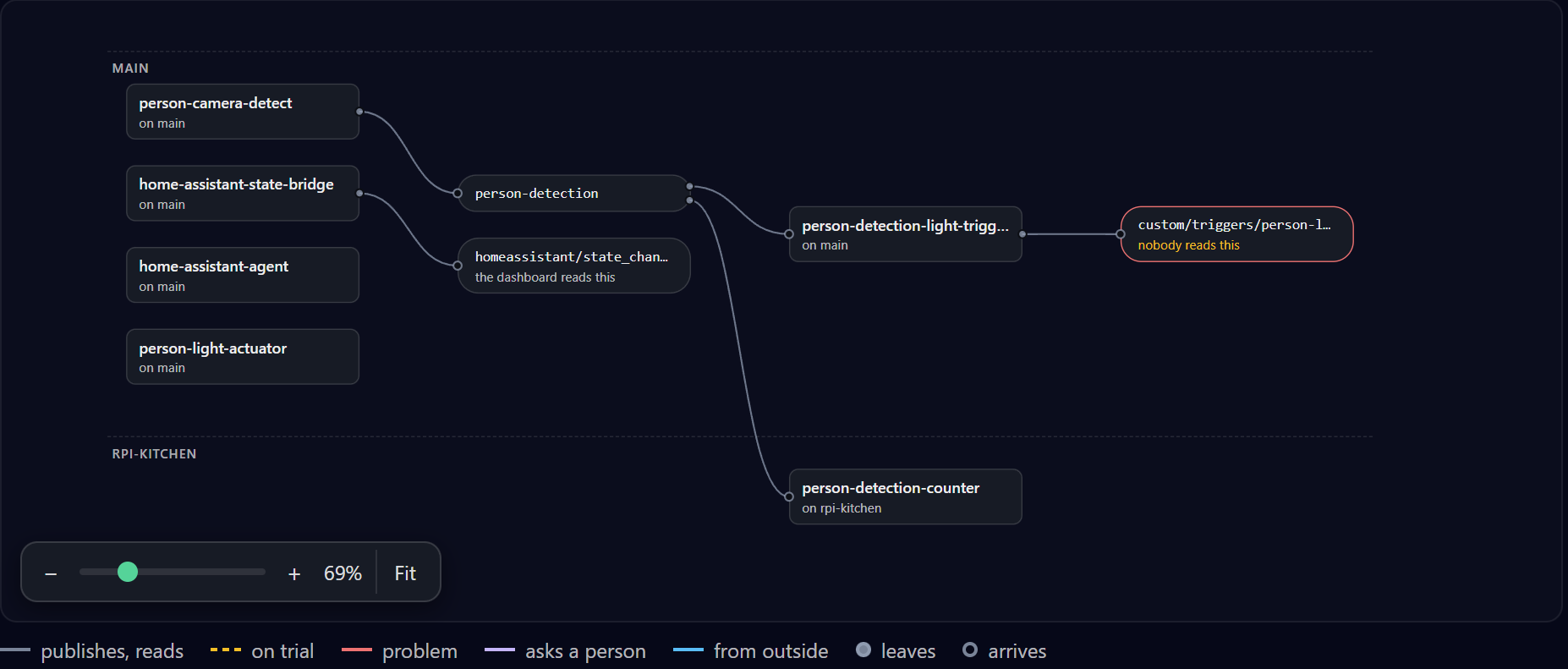}
\caption{The Wiring view during a run of the demonstration on Wactorz~0.8,
grouped by machine. Agents are on the left of each band and topics are
rounded boxes. The detector on the main machine publishes
\texttt{person-detection}, which feeds a lamp trigger on main and a counter
that has been migrated to the \emph{rpi-kitchen} node: the edge crosses
machines because the counter subscribes to a topic, not to an agent. The
trigger's output topic is outlined in red because no agent reads it; a topic
only the dashboard reads is marked as such rather than as a problem.}
\label{fig:wiring}
\end{figure*}

The registry is also what a person needs to see what the system has become
after a few weeks of requests. The dashboard's \emph{Wiring} view
(Figure~\ref{fig:wiring}, also served as \texttt{GET /api/wiring}) draws it as
a graph of agents and topics, optionally grouped by machine, and flags three
kinds of problem: a topic an agent publishes that no agent reads; a
subscriber whose topic has no producer and on which nothing has been seen
(outside the roots where sensors, Home Assistant or scripts are expected to
publish); and a field mismatch, where an agent declares a field it reads that
the topic does not carry. What a topic carries is what was actually observed
on it, falling back to what its producer declares. Selecting an agent or a
topic shows its last payload. The view keeps no state of its own: it is the
topic registry, read at the time of asking.

\subsection{Supervision and self-healing}
Because the code was written by a model, faults are expected rather than
exceptional. Generated code passes a pre-execution validator that blocks
shell execution, file deletion and \texttt{eval}; a sanitiser fixes the most
common LLM mistake (awaiting synchronous methods); repeated runtime errors in
\texttt{setup()}, \texttt{process()} or an MQTT subscribe callback hand the
code and traceback back to the LLM for an in-place repair, which is persisted to
the spawn registry; and \texttt{process()} calls are timeout-guarded. Above that, the
\texttt{Supervisor} applies a ONE\_FOR\_ONE strategy: a crashing agent is
restarted alone and reloads its persisted state. Restarts back off rather
than stop: after a run of consecutive crashes the delay grows to minutes and
then up to an hour, because an outage of something an agent depends on
usually ends, and a retired agent would stay down until someone noticed.
\texttt{MonitorActor} tracks heartbeats and an error registry, restarting
\emph{degraded} agents but deliberately \emph{not} restarting agents whose
code fails to compile (those are reported to the user instead). The spawn
registry itself is persisted, so every dynamic agent comes back after a
reboot.

\subsection{The operator in the loop}
Long-lived agents occasionally need a person. An agent can write to the
operator first, with \texttt{notify\_user}, on the dashboard, Telegram,
Discord or WhatsApp, optionally with options to pick; with
\texttt{ask\_operator} it waits for the answer, and the wait survives
restarts. The dashboard's \emph{Inbox} lists every question an agent is
currently waiting on, alongside the package approvals mentioned above.

\subsection{Edge nodes and live migration}
Any machine with Python~3.11 becomes a node by running the same package as
\texttt{wactorz --node}. \texttt{/deploy~<node>} provisions one over SSH from
a target configured on the main machine (credentials never pass through
chat): it installs Wactorz at main's own version into a virtual environment,
writes the node's configuration, starts it under a systemd unit with a
watchdog, and reports success only after the node has reached the broker and
sent a heartbeat. Main publishes spawn commands to
\texttt{nodes/\{name\}/\allowbreak spawn}; the node compiles and supervises
the agent locally and heartbeats back, so remote agents appear in the same
dashboard.

\texttt{/migrate~<agent>~<node>} moves a \emph{running} agent between main
and a node, or between nodes. Every move goes through main. The agent is
stopped first, so that what its shutdown hook writes last goes with it; its
state travels as JSON, and its blobs (byte strings, \texttt{numpy} arrays,
\texttt{torch} tensors, \texttt{scikit-learn}-family models) travel beside it
in chunks on topics of their own, each checked against its SHA-256 on
arrival. The source keeps its copy until the destination confirms that the
agent started; a destination that never confirms is told to drop it and the
agent is restarted where it was. An agent leaving a node is always restarted
from main's own copy of its program, never from code the node hands back.

\subsection{Security model}
Spawned agents execute code, so the defaults are conservative. The server
binds to loopback unless given an API key. Commands to a node are signed;
the node's link to the broker uses TLS; each node has its own broker account,
limited by an access list; and main accepts a reply, manifest or heartbeat
only from the node it belongs to, so one node cannot answer for, or delete,
another node's agents. Model-written agents can be confined to a
\emph{sandbox} node: in the Docker Compose stack, a container holding none
of main's credentials, whose network reaches the broker and nothing else.

\subsection{Model-agnostic, offline-capable}
Providers (Anthropic, OpenAI, Gemini, NVIDIA~NIM, Ollama) are configured per
call site, so cheap high-frequency calls (intent routing) can run on a local
4B model while code generation uses a hosted one. A built-in evaluation
harness scores models per call site against the production system prompts:
exact-match for routing and Home Assistant classification, \emph{action-set}
match for actuation (an unrequested actuation counts as failure, and some
cases expect a refusal because the device does not exist), a valid plan with
allowed agent types for the planner, and compile-and-define checks for
generated code. It reports accuracy, latency and cost, so that the split is
chosen from measurement rather than guesswork. A deterministic \emph{fake}
provider, and a \texttt{--minimal} profile that runs agents and the
dashboard with no model at all, cover offline use.

\section{Wactorz as a Library}

Everything above the LLM is also available to code a person writes. The
decorator in Listing~\ref{lst:agent} turns a plain function into a supervised
agent: it is called once per message on the subscribed topics, with the
decoded payload, and what it returns is published. It remains an ordinary
function that a test can call directly. A plain function runs on a worker
thread, so a slow model never blocks the event loop the other agents share;
a coroutine runs on the loop.

\begin{listing}[t]
\begin{lstlisting}[language=Python]
import wactorz

@wactorz.agent(
    name="imu-anomaly",
    subscribes="sensors/imu/#",
    publishes="anomalies/imu",
    requires={"packages": ["numpy"]},
    models={"model": "imu-detector:3"})
async def detect(reading, me):
    model = await me.load_model("model")
    score = model_score(model, reading)
    if score < 4.0:
        return None          # publish nothing
    me.persist("anomalies_total",
               int(me.recall("anomalies_total", 0)) + 1)
    return {"score": score, "reading": reading}

wactorz.run(agents=[detect], minimal=True)
\end{lstlisting}
\caption{A hand-written agent: supervised, persistent, wired by topic, and
movable to a node with its code, packages and model.}
\label{lst:agent}
\end{listing}

\texttt{wactorz.run(..., minimal=True)} starts the monitor, the dashboard and
the given agents with no orchestrator, catalogue or installer, so it needs no
model, API key or provider SDK. \texttt{await wactorz.serve(...)} does the
same inside a program that already has an event loop: a Jupyter notebook, a
FastAPI application or a ROS node. Agents with their own lifecycle subclass
\texttt{Actor}, whose class attributes (\texttt{SUBSCRIBES},
\texttt{PUBLISHES}, input and output schemas) become the agent's catalogue
entry and the contracts a pipeline is checked against. Steps can be declared
together as a \emph{pipeline} with a schedule and typed rules, and the
planner can wire newly generated agents to what hand-written ones publish.

A registered library agent migrates like a generated one. Main sends the
folder its module is in (Python files only) along with the spawn; the node
checks it against the hash the signed spawn names, installs the packages
the agent declares, fetches the models it names, and builds the agent.
Imports from outside that folder, other than the standard library, Wactorz
and declared packages, are refused before anything stops, naming the import.
A node makes no model calls, so an agent's \texttt{me.llm} is \texttt{None}
there.

Wactorz does not replace LangGraph or AG2; it keeps them running. An
\texttt{async} agent that invokes a compiled graph is the whole integration,
and the graph's nodes can call the system's model through \texttt{me.llm},
so their spend is counted with every other call. The chat itself is answered
by an \emph{orchestrator}, a three-method protocol: main is the
LLM-driven default, and a deployment can supply its own, a LangGraph graph or
a rule engine, without changing anything else.

\section{Models at the Edge}

Physical-AI agents usually serve a trained model, and the model changes more
often than the agent's code. Wactorz keeps models on the main machine, by name
and version, with \texttt{wactorz-artifacts}; each version resolves to the
SHA-256 of its bytes, and identical bytes are stored once. An agent names the
models it needs (\texttt{models=} in Listing~\ref{lst:agent}, or an
\texttt{artifacts} entry in a generated agent's spawn configuration) and opens
them with \texttt{load\_model}, which reads \texttt{npz}, \texttt{npy},
\texttt{safetensors}, \texttt{state\_dict} checkpoints and ONNX in ways that
execute no code from the file. Pickles, \texttt{joblib} files and whole
\texttt{torch} modules, which do run code when loaded, are refused unless the
caller passes \texttt{allow\_code=True}; other formats (GGUF, TFLite) are
opened by path.

\paragraph{Transfer to nodes.} When an agent that names a model is placed on
a node, main pins each reference to the exact version and bytes it means at
that moment, so the agent restarted or moved later runs the same model. The
node asks main only for what its cache lacks, checks each file against its
hash, and starts the agent only once every file is present; a migration whose
model does not arrive is put back where it was. Main sends a node only the
models of the agents placed on it.

\paragraph{Replacing a model.} Two pointers per model say which version is
\emph{active} (served) and which is \emph{shadow}: on trial beside it, on the
same inputs, acted on by nobody. Every pointer move records who made it and
why, and is published on a retained topic so every machine hears it. A
version can declare which label values count as what it should flag; a
trial then judges both versions against labels that operators supply,
reporting precision, recall, F1, false alarms per day and time to detection
with confidence intervals and a test of whether the two differ. A
\emph{promotion policy} (metric, margin, minimum days and labelled inputs)
decides when the shadow may replace the active version, automatically or after
the operator confirms on the dashboard. A \texttt{trainer} agent can produce
versions inside the loop, from a user-written recipe run on a window of
recorded readings in a child process; and a \texttt{drift} agent compares
recent readings with the window the served version was trained on (PSI, a
two-sample KS test, and ADWIN on the model's scores). An alarm is an ordinary
topic, so a rule can send the trainer a task, and the policy decides whether
the result is promoted: the loop closes without a person, unless the policy
asks for one.

\section{Demonstration}

The station runs Wactorz on a laptop with its webcam, a Home-Assistant-controlled
lamp and a Raspberry~Pi; the accompanying video is one unedited run (Claude
Sonnet~5 at every call site; total cost \$0.13). Visitors type into the chat and
watch the Feed:

\begin{enumerate}
\item \textbf{Create.} \emph{``Spawn an agent to detect if there is any person;
  if there is, turn on the living-room light. Publish every detection to
  \texttt{camera/livingroom/person} \ldots use my laptop's camera.''} The planner
  proposes a YOLO detector (with \texttt{ultralytics} and OpenCV declared as
  installs) and a Home Assistant actuator; on
  \emph{yes}, both spawn and the lamp lights within seconds.
\item \textbf{Extend.} \emph{``Add an agent that counts when a person is
  detected. Persist the running total. When the counter is $\geq 10$, divide it
  by zero---my system can handle such errors.''} The counter is auto-wired to
  the detector's topic and starts incrementing; the Wiring view shows the new
  edge.
\item \textbf{Break and heal.} At count~10 the callback raises a
  \texttt{Zero\-Division\-Error} three times in four seconds; the Feed shows
  \emph{Runtime error---asking LLM to fix}, then \emph{has recovered and is
  running normally again}. The repaired code no longer divides; the count
  continues 10, 11, 12 without a reset, and the lamp keeps actuating throughout.
\item \textbf{Deploy.} \texttt{/deploy rpi-kitchen} provisions a bare
  Raspberry~Pi over SSH (virtual environment, Wactorz, systemd unit) while the
  counter keeps ticking; the node appears in the dashboard.
\item \textbf{Move.} \texttt{/migrate livingroom-\allowbreak person-counter rpi-kitchen}
  stops the local instance and re-spawns it on the Pi with its state: the count
  continues from the new node, \texttt{/nodes} reports
  \emph{rpi-kitchen $\cdot$ 1 agent}, and the Wiring view, grouped by
  machine, shows the counter under \emph{rpi-kitchen}, still reading the
  detector's topic on main (Figure~\ref{fig:wiring}).
\end{enumerate}
Without network access, every call site can run on a local model through an Ollama-compatible interface.

\section{Limitations}

\paragraph{The validator is not a sandbox.} The pre-execution check is a
blocklist over the code's syntax tree. A name it cannot follow (held in a
variable, looked up in \texttt{sys.modules}) passes, and so does anything a
permitted module can do, such as deleting a file through \texttt{pathlib}.
Inside one process, all agents share one privilege level: there is no
isolation between them. Isolation comes from placement instead, by running
model-written code on the sandbox node or on a separate machine.

\paragraph{A repair restores liveness, not intent.} In the demonstration the
repaired counter no longer divides by zero, which is what keeps it running,
but the user had asked for the division. The LLM repair optimises for code
that runs; whether it still does what was asked is for the operator to
judge, and the repaired source is persisted and visible for that reason.

\paragraph{Migration has limits.} State that can be neither serialised nor
rebuilt (an open camera handle, a socket, an object of the agent's own class)
blocks a move unless it is forced, in which case those keys are left behind
and named. State and blobs are capped (8\,MiB and 64\,MiB by default), and
values that run code when they load (a pickled model, a whole
\texttt{torch} module) are sent from main to a node but never accepted back
from one, so an agent that must return keeps a \texttt{state\_dict} instead.
Main and its nodes must run the same release series; a node is redeployed
when main is upgraded.

\paragraph{Evaluation.} The demonstration shows one run of one scenario; we
report no systematic measurements of repair success rates across models or
fault types. The evaluation harness is the instrument for such a study, which
we leave to future work.

\section{Significance}
Wactorz bridges the LLM-agent, distributed-systems and IoT communities: it
shows that classical supervision makes LLM-written code safe enough to run
unattended, that pub/sub is a sufficient substrate for spawning and migrating
agents, and the models they serve, across heterogeneous hardware, and that
the whole stack can be self-hosted. The same runtime hosts hand-written
agents, so a team can mix code it wrote, code a model wrote, and graphs from
other frameworks under one supervisor and one dashboard. It ships as a
\texttt{pip} package, Docker image and Home Assistant add-on under
Apache-2.0, with a 500-file test suite and an extensible evaluation harness.

\bibliography{refs}

\end{document}